\def\year{2017}\relax
\documentclass[letterpaper]{article}
\usepackage[hyperindex,breaklinks,draft]{hyperref}
\usepackage{aaai17}
\nocopyright
\usepackage{times}
\usepackage{helvet}
\usepackage{courier}
\usepackage{amsmath,amssymb}
\usepackage{booktabs}
\usepackage{bm}
\usepackage{color}
\usepackage{graphicx}
\usepackage{url}
\usepackage{breqn}
\usepackage{empheq}
\usepackage{booktabs}
\usepackage[noend]{algpseudocode}
\usepackage{algorithm}

\newif\ifaaai
\aaaifalse

\usepackage{notation}
\newcommand{\todo}[1]{[\textcolor{red}{TODO: #1}]}
\renewcommand{\emph}[1]{\textbf{#1}}
\newcommand{\sys}[1]{\textsc{#1}}

\newcommand{\mycite}[1]{\citeauthor{#1} \shortcite{#1}}
\newcommand{\nb}{Na\"ive Bayes}
\def\thetitle{Unsupervised Learning for Lexicon-Based Classification}
\def\theauthor{Jacob Eisenstein}
\begin{document}
%
\title{\thetitle}
\author{\theauthor\\
Georgia Institute of Technology\\
801 Atlantic Drive NW\\
Atlanta, Georgia 30318\\
}
\maketitle
\begin{abstract}
In lexicon-based classification, documents are assigned labels by  comparing the number of words that appear from two opposed lexicons, such as positive and negative sentiment. Creating such words lists is often easier than labeling instances, and they can be debugged by non-experts if classification performance is unsatisfactory. However, there is little analysis or justification of this classification heuristic. This paper describes a set of assumptions that can be used to derive a probabilistic justification for lexicon-based classification, as well as an analysis of its expected accuracy. One key assumption behind lexicon-based classification is that all words in each lexicon are equally predictive. This is rarely true in practice, which is why lexicon-based approaches are usually outperformed by supervised classifiers that learn distinct weights on each word from labeled instances. This paper shows that it is possible to learn such weights without labeled data, by leveraging co-occurrence statistics across the lexicons. This offers the best of both worlds: light supervision in the form of lexicons, and data-driven classification with higher accuracy than traditional word-counting heuristics.
\end{abstract}

\section{Introduction}
\emph{Lexicon-based classification} refers to a classification rule in which documents are assigned labels based on the count of words from lexicons associated with each label~\cite{taboada2011lexicon}. For example, suppose that we have opposed labels $Y \in \{0,1\}$, and we have associated lexicons $\lex_0$ and $\lex_1$. Then for a document with a vector of word counts $\vx$, the lexicon-based decision rule is,
\begin{dmath}
\label{eq:lexicon-decision-rule}
\sum_{i \in \lex_0} x_i \gtrless \sum_{j \in \lex_1} x_j,
\end{dmath}
where the $\gtrless$ operator indicates a decision rule. Put simply, the rule is to select the label whose lexicon matches the most word tokens.

Lexicon-based classification is widely used in industry and academia, with applications ranging from sentiment classification and opinion mining~\cite{pang2008opinion,liu2015sentiment} to the psychological and ideological analysis of texts~\cite{laver2000estimating,tausczik2010psychological}. 
The popularity of this approach can be explained by its relative simplicity and ease of use: for domain experts, creating lexicons is intuitive, and, in comparison with labeling instances, it may offer a faster path towards a reasonably accurate classifier~\cite{settles2011closing}. Furthermore, classification errors can be iteratively debugged by refining the lexicons.

However, from a machine learning perspective, there are a number of drawbacks to lexicon-based classification. First, while intuitively reasonable, lexicon-based classification lacks theoretical justification: it is not clear what conditions are necessary for it to work. Second, the lexicons may be incomplete, even for designers with strong substantive intuitions. Third, lexicon-based classification assigns an equal weight to each word, but some words may be more strongly predictive than others.\footnote{Some lexicons attach coarse-grained predefined weights to each word. For example, the OpinionFinder Subjectivity lexicon labels words as ``strongly'' or ``weakly'' subjective~\cite{wilson2005recognizing}. This poses an additional burden on the lexicon creator.}
Fourth, lexicon-based classification ignores multi-word phenomena, such as negation (e.g., \ex{not so good}) and discourse (e.g., \ex{the movie would be watchable if it had better acting}). Supervised classification systems, which are trained on labeled examples, tend to outperform lexicon-based classifiers, even without accounting for multi-word phenomena~\cite{liu2015sentiment,pang2008opinion}.

Several researchers have proposed methods for \emph{lexicon expansion}, automatically growing lexicons from an initial seed set~\cite{hatzivassiloglou1997predicting,qiu2011opinion}. There is also work on handling multi-word phenomena such as negation~\cite{wilson2005recognizing,polanyi2006contextual}, and discourse~\cite{somasundaran2008discourse,bhatia2015better}. However, the theoretical foundations of lexicon-based classification remain poorly understood, and we lack principled means for automatically assigning weights to lexicon items without resorting to labeled instances. 

This paper elaborates a set of assumptions under which lexicon-based classification is equivalent to \nb\ classification. I then derive expected error rates under these assumptions. These expected error rates are not matched by observations on real data, suggesting that the underlying assumptions are invalid. Of key importance is the assumption that each lexicon item is equally predictive. To relax this assumption, I derive a principled method for estimating word probabilities under each label, using a method-of-moments estimator on cross-lexical co-occurrence counts. 

Overall, this paper makes the following contributions:
\begin{itemize}
\item justifying lexicon-based classification as a special case of multinomial \nb;
\item mathematically analyzing this model to compute the expected performance of lexicon-based classifiers;
\item extending the model to justify a popular variant of lexicon-based classification, which incorporates word presence rather than raw counts;
\item deriving a method-of-moments estimator for the parameters of this model, enabling lexicon-based classification with unique weights per word, without labeled data;
\item empirically demonstrating that this classifier outperforms lexicon-based classification and alternative approaches.
\end{itemize}

\section{Lexicon-Based Classification as \nb}
I begin by showing how the lexicon-based classification rule shown in (\ref{eq:lexicon-decision-rule}) can be derived as a special case of \nb\ classification. Suppose we have a prior probability $P_Y$ for the label $Y$, and a likelihood function $P_{X \mid Y}$, where $X$ is a random variable corresponding to a vector of word counts. The conditional label probability can be computed by Bayesian inversion,
\begin{dmath}
P(y \mid \vx) = \frac{P(\vx \mid y) P(y)}{\sum_{y'} P(\vx \mid y')P(y')}.
\end{dmath}

Assuming that the costs for each type of misclassification error are identical, then the minimum Bayes risk classification rule is,
\begin{dmath}
  \label{eq:basic-decision-rule}
  \log \Pr(Y=0) \hiderel{+} \log P(\vx \mid Y=0)\\ \gtrless \log \Pr(Y=1) \hiderel{+} \log P(\vx \mid Y=1),
\end{dmath}
moving to the log domain for simplicity of notation. I now show that lexicon-based classification can be justified under this decision rule, given a set of assumptions about the probability distributions. 

Let us introduce some assumptions about the likelihood function, $P_{X \mid Y}$. The random variable $X$ is defined over vectors of counts, so a natural choice for the form of this likelihood is the multinomial distribution, corresponding to a multinomial \nb\ classifier. For a specific vector of counts $X = \vx$, write $P(\vx~\mid~y) \triangleq P_{\text{multinomial}}(\vx ; \vth_y, N)$, where $\vth_y$ is a probability vector associated with label $y$, and $N = \sum^V_{i=1} x_i$ is the total count of tokens in $\vx$, and $x_i$ is the count of word $i \in \{1, 2, \ldots, V\}$. The multinomial likelihood is proportional to a product of likelihoods of categorical variables corresponding to individual words (tokens), 
\begin{equation}
\Pr(W=i \mid Y=y; \vth) = \theta_{y,i},
\end{equation}
where the random variable $W$ corresponds to a single token, whose probability of being word type $i$ is equal to $\theta_{y,i}$ in a document with label $y$. The multinomial log-likelihood can be written as,

\begin{align}
\notag
\log P(\vx \mid y) = & \log \Pm(\vx; \vth_y, N)\\
\notag
= & K(\vx) + \sum^V_{i=1} x_i \log \Pr(W= i \mid Y = y; \vth)\\
= & K(\vx) + \sum^V_{i=1} x_i \log \theta_{y,i},
\end{align}
where $K(\vx)$ is a function of $\vx$ that is constant in $y$. 

The first necessary assumption about the likelihood function is that the lexicons are \emph{complete}: words that are in neither lexicon have identical probability under both labels. Formally, for any word $i \notin \lex_0 \cup \lex_1$, we assume,
\begin{equation}
\Pr(W = i \mid Y=0) = \Pr(W = i \mid Y=1),
\label{eq:assume-complete}
\end{equation}
which implies that these words are irrelevant to the classification boundary.

Next, we must assume that each in-lexicon word is \emph{equally predictive}. Specifically, for words that are in lexicon $y$,
\begin{equation}
\frac{\Pr(W = i \mid Y = y)}{\Pr(W = i \mid Y = \neg y)} = \frac{1+\gamma}{1-\gamma},
\label{eq:assume-equal}
\end{equation}
where $\neg y$ is the opposite label from $y$. The parameter $\gamma$ controls the predictiveness of the lexicon: for example, if $\gamma =0.5$ in a sentiment classification problem, this would indicate that words in the positive sentiment lexicon are three times more likely to appear in documents with positive sentiment than in documents with negative sentiment, and vice versa. The word \ex{atrocious} might be less likely overall than \ex{good}, but still three times more likely in the negative class than in the positive class. In the limit, $\gamma=0$ implies that the lexicons do not distinguish the classes at all, and $\gamma=1$ implies that the lexicons distinguish the classes perfectly, so that the observation of a single in-lexicon word would completely determine the document label.

The conditions enumerated in (\ref{eq:assume-complete}) and (\ref{eq:assume-equal}) are ensured by the following definition,
\begin{equation}
\theta_{y,i} = \begin{cases}
(1+\gamma) \mu_i, & i \in \lex_y \\
(1-\gamma) \mu_i, & i \in \lex_{\neg y}\\
\mu_i, & i \notin \lex_y \cup \lex_{\neg y},
\end{cases}
\label{eq:def-theta}
\end{equation}
where $\neg y$ is the opposite label from $y$, and $\vmu$ is a vector of baseline probabilities, which are independent of the label.

Because the probability vectors $\vth_0$ and $\vth_1$ must each sum to one, we require an assumption of \emph{equal coverage}, \begin{equation}
\sum_{i \in \lex_0} \mu_i = \sum_{j \in \lex_1} \mu_j.
\end{equation}

Finally, assume that the labels have \emph{equal prior likelihood}, $\Pr(Y=0) = \Pr(Y=1)$. It is trivial to relax this assumption by adding a constant term to one side of the decision rule in (\ref{eq:lexicon-decision-rule}).

With these assumptions in hand, it is now possible to simplify the decision rule in (\ref{eq:basic-decision-rule}). Thanks to the assumption of equal prior probability, we can drop the priors $P(Y)$, so that the decision rule is a comparison of the likelihoods,
\begin{align}
\log P(\vx \mid Y=0) & \gtrless \log P(\vx \mid Y=1)\\
K(\vx) + \sum_i x_i \log \theta_{0,i} & \gtrless K(\vx) + \sum_i x_i \log \theta_{1,i}.
\end{align}
Canceling $K(\vx)$ and applying the definition from (\ref{eq:def-theta}),
\begin{dmath}
 \sum_{i \in \lex_0} x_i \log ((1+\gamma)\mu_i) + \sum_{i \in \lex_1} x_i \log ((1-\gamma)\mu_i)\\
  \gtrless \sum_{i \in \lex_0} x_i \log ((1-\gamma)\mu_i) + \sum_{i \in \lex_1} x_i \log ((1+\gamma)\mu_i).
\end{dmath}
The $\mu_i$ terms cancel after distributing the $\log$, leaving,
\begin{align}
\sum_{i \in \lex_0} x_i \log \frac{1+\gamma}{1-\gamma}
\gtrless \sum_{i \in \lex_1} x_i \log \frac{1+\gamma}{1-\gamma}.
\label{eq:derived-rule}
\end{align}
For any $\gamma \in (0,1)$, the term $\log \frac{1+\gamma}{1-\gamma}$ is a finite and positive constant. Therefore, (\ref{eq:derived-rule}) is identical to the counting-based classification rule in (\ref{eq:lexicon-decision-rule}). In other words, lexicon-based classification is minimum Bayes risk classification in a multinomial probability model, under the assumptions of equal prior likelihood, lexicon completeness, equal predictiveness of words, and equal coverage.

\section{Analysis of Lexicon-Based Classification}
One advantage of deriving a formal foundation for lexicon-based classification is that it is possible to analyze its expected performance. For a label $y$, let us write the count of in-lexicon words as $m_y = \sum_{i \in \lex_y} x_i$, and the count of opposite-lexicon words as $m_{\neg y} = \sum_{i \in \lex_{\neg y}} x_{i}$. Lexicon-based classification makes a correct prediction whenever $m_y > m_{\neg y}$ for the correct label $y$. To assess the likelihood that $m_y > m_{\neg y}$, it is sufficient to compute the expectation and variance of the difference $m_y - m_{\neg y}$; under the central limit theorem, we can treat this difference as approximately normally distributed, and compute the probability that the difference is positive using the Gaussian cumulative distribution function (CDF).

\newcommand\musum{\ensuremath{s_{\mu}}}
Let us use the convenience notation $\musum$,
\begin{equation}
\musum \triangleq \sum_{i \in \lex_{0}} \mu_i = \sum_{i \in \lex_1}\mu_i.
\end{equation}
Recall that we have already taken the assumption that the sums of baseline word probabilities for the two lexicons are equal. Under the multinomial probability model, given a document with $N$ tokens, the expected counts are,
\begin{align}
E[m_y] = & N \sum_{i \in \lex_{y}}\theta_{y,i} = N(1+\gamma) \musum \\
  E[m_{\neg y}] = & N \sum_{i \in \lex_{\neg y}}\theta_{\neg y,i} = N(1-\gamma) \musum \\
E[m_y - m_{\neg y}] = & 2N\gamma \musum.
\end{align}

Next we compute the variance of this margin,
\begin{align}
V[m_y - m_{\neg y}] = & V[m_y] + V[m_{\neg y}] + Cov(m_y, m_{\neg y}).
\end{align}
Each of these terms is the variance of a sum of counts. Under the multinomial distribution, the variance of a single count is $V[x_i] = N \theta_i (1-\theta_i)$. The variance of the sum $m_y$ is,
\begin{align}
\notag
V[m_y] = & \sum_{i \in \lex_{y}} N \theta_i (1-\theta_i) - \sum_{j \in \lex_{y}, j \neq i} N \theta_i \theta_j\\
= & \sum_{i \in \lex_{y}} N \theta_i - N \theta_i^2 - \sum_{j \in \lex_{y}, j \neq i} N \theta_i \theta_j\\
\notag
\leq & N \sum_{i \in \lex_{y}} \theta_i = N \sum_{i \in \lex_{y}} (1+\gamma)\mu_i\\
= & N(1+\gamma) \musum.
\end{align}

An equivalent upper bound can be computed for the variance of the count of opposite lexicon words, 
\begin{equation}
V[m_{\neg y}] \leq N(1-\gamma)\musum.
\end{equation}
These bounds are fairly tight because the products of probabilities $\theta_i^2$ and $\theta_i \theta_j$ are nearly always small, due to the fact that most words are rare. Because the covariance $Cov(m_y,m_{\neg y})$ is negative (and also involves a product of word probabilities), we can further bound the variance of the margin, obtaining the upper bound,
\begin{equation}
V[m_y - m_{\neg y}] \leq N (1+\gamma)\musum + N(1-\gamma)\musum
= 2 N \musum.
\end{equation}

By the central limit theorem, the margin $m_y - m_{\neg y}$ is approximately normally distributed, with mean $2N\gamma \musum$ and variance upper-bounded by $2N\musum$. The probability of making a correct prediction (which occurs when $m_y > m_{\neg y}$) is then equal to the cumulative density of a standard normal distribution $\Phi(z)$, where the $z$-score is equal to the ratio of the expectation and the standard deviation,
\begin{align}
z = \frac{E[m_y - m_{\neg y}]}{\sqrt{V[m_y - m_{\neg y}]}} 
\geq \frac{2N\gamma \musum}{\sqrt{2N\musum}}
= \gamma \sqrt{2N\musum}.
\end{align}
Note that by upper-bounding the variance, we obtain a lower bound on the $z$-score, and thus a lower bound on the expected accuracy.

According to this approximation, accuracy is expected to increase with the predictiveness $\gamma$, the document length $N$, and the lexicon coverage $\musum$. This helps to explain a dilemma in lexicon design: as more words are added, the coverage increases, but the average predictiveness of each word decreases (assuming the most predictive words are added first). Thus, increasing the size of a lexicon by adding marginal words may not improve performance. 

The analysis also predicts that longer documents should be easier to classify. This is because the expected size of the gap $m_y - m_{\neg y}$ grows with $N$, while its standard deviation grows only with $\sqrt{N}$. This prediction can be tested empirically, and on all four datasets considered in this paper, it is false: longer documents are harder to classify accurately. This is a clue that the underlying assumptions are not valid. The decreased accuracy for especially long reviews may be due to these reviews being more complex, perhaps requiring modeling of the discourse structure~\cite{somasundaran2008discourse}.

\section{Justifying the Word-Appearance Heuristic}
An alternative heuristic to lexicon-based classification is to consider only the \emph{presence} of each word type, and not its count. This corresponds to the decision rule,
\begin{dmath}
\label{eq:appearance-decision-rule}
\sum_{i \in \lex_0} \delta(x_i > 0) \gtrless \sum_{j \in \lex_1} \delta(x_j > 0),
\end{dmath}
where $\delta(\cdot)$ is a delta function that returns one if the Boolean condition is true, and zero otherwise. In the context of supervised classification, \mycite{pang2002thumbs} find that word presence is a more predictive feature than word frequency. By ignoring repeated mentions of the same word, heuristic (\ref{eq:appearance-decision-rule}) emphasizes the diversity of ways in which a document covers a lexicon, and is more robust to document-specific idiosyncrasies --- such as a review of \ex{The Joy Luck Club}, which might include the positive words \ex{joy} and \ex{luck} many times even if the review is negative.

The word-appearance heuristic can also be explained in the framework defined above. The multinomial likelihood $P_{X \mid Y}$ can be replaced by a \emph{Dirichlet-compound multinomial} (DCM) distribution, also known as a multivariate Polya distribution~\cite{madsen2005modeling}. This distribution is written $\Pdcm(\vx; \va_y)$, where $\va_y$ is a vector of parameters associated with label $y$, with $\alpha_{y,i} > 0$ for all $i \in \{1, 2, \ldots, V\}$. The DCM is a ``compound'' distribution because it treats the parameter of the multinomial as a latent variable to be marginalized out,
\begin{small}
\begin{equation}
\Pdcm(\vx; \vec{\alpha}_y) = \int_{\vec{\nu}} \Pm(\vx \mid \vec{\nu}) \Pd(\vec{\nu} \mid \va_y) d\vec{\nu}.
\end{equation}
\end{small}
Intuitively, one can think of the DCM distribution as encoding a model in which each document has its own multinomial distribution over words; this document-specific distribution is itself drawn from a prior that depends on the class label $y$.

Suppose we set the DCM parameter $\va = \conc \vth$, with $\vth$ as defined in (\ref{eq:def-theta}). The constant $\conc > 0$ is then the \emph{concentration} of the distribution: as $\conc$ grows, the probability distribution over $\va$ is more closely concentrated around the prior expectation $\vth$. Because $\sum^V_i \theta_i = 1$, the likelihood function under this model is,
\begin{equation}
\Pdcm(\vx \mid y) = 
\frac{\Gamma(\conc)}{\Gamma(N + \tau)}
\prod_i \frac{\Gamma(x_i + \conc \theta_{y,i})}{\Gamma(\conc \theta_{y,i})},
\end{equation}
where $\Gamma(\cdot)$ is the gamma function. Minimum Bayes risk classification in this model implies the decision rule:
\begin{align}
\sum_{i \in \lex_0} \log \frac{r_{\text{in}}(x_{i})}{r_{\text{out}}(x_{i})} \gtrless &
\sum_{i \in \lex_1} \log \frac{r_{\text{in}}(x_{t,i})}{r_{\text{out}}(x_{i})}
\label{eq:dcm-rule}
\end{align}
where,
\begin{align}
\label{eq:r-in}
r_{\text{in}}(x_i) \triangleq & 
\frac{\Gamma(x_{i} + \conc (1+\gamma) \mu_i)}
{\Gamma(\conc (1+\gamma) \mu_i)}\\
\label{eq:r-out}
r_{\text{out}}(x_{i}) \triangleq & 
\frac{\Gamma(x_{i} + \conc (1-\gamma) \mu_i)}
{\Gamma(\conc (1-\gamma) \mu_i)}.
\end{align}

\begin{figure}
\centering
\includegraphics[width=0.39\textwidth]{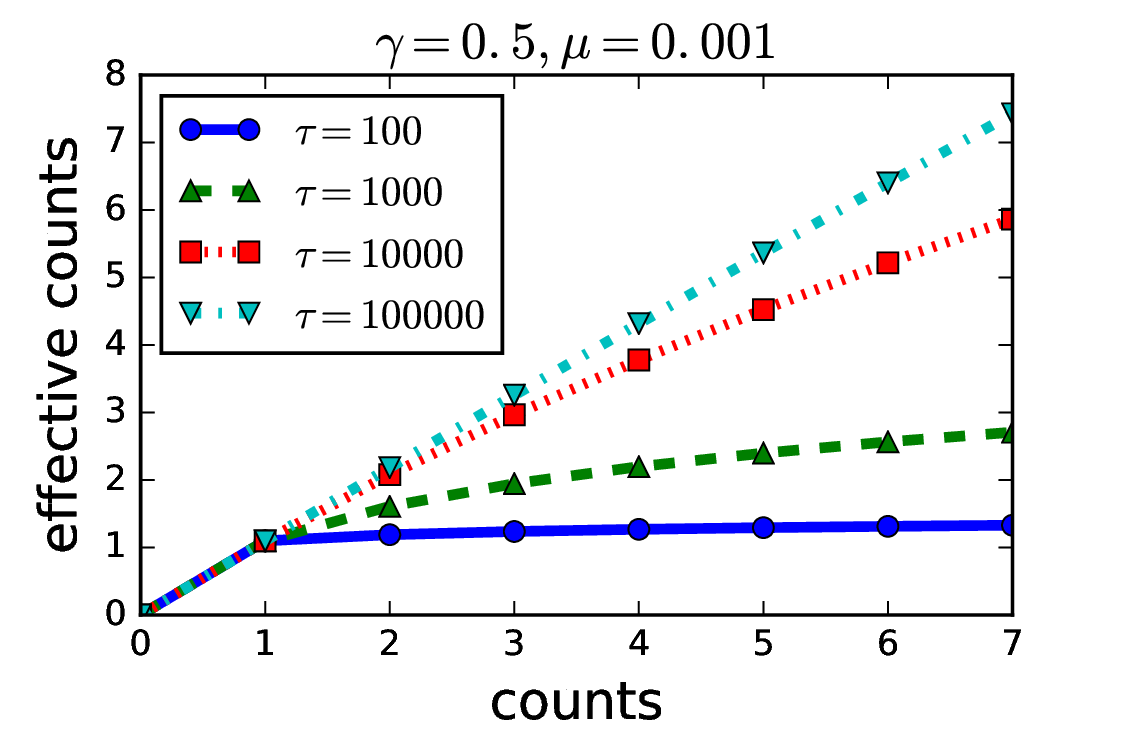}
\caption{Effective counts for varying values of $\conc$. For the datasets considered in this paper, $\tau$ usually falls in the range between $500$ and $1000$.}
\label{fig:effective-counts}
\end{figure}

As $\conc \to \infty$, the prior on $\vec{\nu}$ is tightly linked to $\vth$, so that the model reduces to the multinomial defined above. Another way to see this is to apply the equality $\Gamma(x+1) = x\Gamma(x)$ to (\ref{eq:r-in}) and (\ref{eq:r-out}) when $\conc\mu_i \gg x_i$. As $\conc \to 0$, the prior on $\vec{\nu}$ becomes increasingly diffuse. Repeated counts of any word are better explained by document-specific variation from the prior, than by properties of the label. This situation is shown in Figure~\ref{fig:effective-counts}, which plots the ``effective counts'' implied by the classification rule (\ref{eq:dcm-rule}) for a range of values of the concentration parameter $\conc$, holding the other parameters constant ($\mu = 10^{-3}, \gamma = 0.5$). For high values of $\conc$, the effective counts track the observed counts linearly, as in the multinomial model; for low values of $\conc$, the effective counts barely increase beyond $1$.

\mycite{minka2000estimating} presents a number of estimators for the concentration parameter $\conc$ from a corpus of text. When the label $y$ is unknown, we cannot apply these estimators directly. However, as described above, out-of-lexicon words are assumed to have identical probability under both labels. This assumption can be exploited to estimate $\conc$ exclusively from the first and second moments of these out-of-lexicon words. Analysis of the expected accuracy of this model is left to future work.

\section{Estimating Word Predictiveness}
A crucial simplification made by lexicon-based classification is that all words in each lexicon are equally predictive. In reality, words may be more or less predictive of class labels, for reasons such as sense ambiguity (e.g., \ex{well}) and degree (e.g., \ex{good} vs \ex{flawless}). By introducing a per-word predictiveness factor $\gamma_i$ into (\ref{eq:def-theta}), we arrive at a model that is a restricted form of Na\"ive Bayes. (The restriction is that the probabilities of non-lexicon words are constrained to be identical across classes.) If labeled data were available, this model could be estimated by maximum likelihood. This section shows how to estimate the model without labeled data, using the method of moments.

First, note that the baseline probabilities $\mu_i$ can be estimated directly from counts on an unlabeled corpus; the challenge is to estimate the parameters $\gamma_i$ for all words in the two lexicons. The key intuition that makes this possible is that highly predictive words should rarely appear with words in the opposite lexicon. This idea can be formalized in terms of \emph{cross-label counts}: the cross-label count $c_i$ is the co-occurrence count of word $i$ with all words in the opposite lexicon,
\begin{equation}
c_i = \sum_{t=1}^T \sum_{j \in \lex_{\neg y}} x^{(t)}_{i} x^{(t)}_{j},
\end{equation}
where $\vx^{(t)}$ is the vector of word counts for document $t$, with $t \in \{1\ldots T\}$. Under the multinomial model defined above, for a single document with $N$ tokens, the expected product of counts for a word pair $(i,j)$ is equal to,
\begin{align}
\notag
E[x_i x_j] = & E[x_i] E[x_j] + Cov(x_i,x_j)\\
\notag
= & N\theta_i N \theta_j - N \theta_i \theta_j\\
= & N(N-1) \theta_i \theta_j.
\end{align}

Let us focus on the expected products of counts for cross-lexicon word pairs $(i \in \lex_0, j \in \lex_1)$. The parameter $\vth$ depends on the document label $y$, as defined in (\ref{eq:def-theta}). As a result, we have the following expectations,
\begin{small}
\begin{align}
\notag
E[x_i x_j \mid Y = 0] = & N(N-1) \mu_i (1 + \gamma_i) \mu_j (1 - \gamma_j) \\
= & N(N-1)\mu_i \mu_j (1 + \gamma_i - \gamma_j - \gamma_i \gamma_j)\\
\notag
E[x_i x_j \mid Y = 1] = & N(N-1) \mu_i (1 - \gamma_i) \mu_j (1 + \gamma_j) \\
= & N(N-1)\mu_i \mu_j (1 - \gamma_i + \gamma_j - \gamma_i \gamma_j)\\
\notag
E[x_i x_j] = & P(Y=0)E[x_i x_j \mid Y = 0] \\
\notag
& + P(Y=1)E[x_i x_j \mid Y = 1]\\
= & N (N-1) \mu_i \mu_j (1 -\gamma_i \gamma_j).
\end{align}
\end{small}
Summing over all words $j \in \lex_1$ and all documents $t$, 
\begin{dmath}
E[c_i] =  \sum^T_{t=1} \sum_{j\in \lex_1} E[x^{(t)}_i x^{(t)}_j]\\
 =  \sum^T_{t=1} N_t (N_t - 1) 
\mu_i \sum_{j\in \lex_1}  \mu_j (1 - \gamma_i \gamma_j) 
\label{eq:expected-co-count}
\end{dmath}

Let us write $\vg^{(1)}$ to indicate the vector of $\gamma_j$ parameters for all $j \in \lex_1$, and $\vg^{(0)}$ for all $i \in \lex_0$. The expectation in (\ref{eq:expected-co-count}) is a linear function of $\gamma_i$, and a linear function of the vector $\vg^{(1)}$. Analogously, for all $j\in\lex_1$, $E[c_j]$ is a linear function of $\gamma_j$ and $\vg^{(1)}$. Our goal is to choose $\vg$ so that the expectations $E[c_i]$ closely match the observed counts $c_i$. This can be viewed as form of \emph{method of moments} estimation, with the following objective, 
\begin{equation}
J = \frac{1}{2} \sum_{i\in \lex_0}(c_i - E[c_i])^2 + \frac{1}{2}\sum_{j \in \lex_1}(c_j - E[c_j])^2,
\end{equation}
which can be minimized in terms of $\vg^{(0)}$ and $\vg^{(1)}$. However, there is an additional constraint: the probability distributions $\vth_0$ and $\vth_1$ must still sum to one. We can express this as a linear constraint on $\vg^{(0)}$ and $\vg^{(1)}$,
\begin{equation}
\vmu^{(0)} \cdot \vg^{(0)} - \vmu^{(1)} \cdot \vg^{(1)} = 0,
\end{equation}
where $\vmu^{(y)}$ is the vector of baseline probabilities for words $i \in \lex_y$, and $\vmu^{(0)} \cdot \vg^{(0)}$ indicates a dot product. 

We therefore formulate the following constrained optimization problem,
\begin{align}
\notag
\min_{\vg^{(0)}, \vg^{(1)}} & \:\: \frac{1}{2} \sum_{i\in \lex_0}(c_i - E[c_i])^2 + \frac{1}{2}\sum_{j \in \lex_1}(c_j - E[c_j])^2\\
\notag
s.t. &  \:\:  \vmu^{(0)} \cdot \vg^{(0)} - \vmu^{(1)} \cdot \vg^{(1)}=0 \\
& \:\: \forall i \in (\lex_0 \cup \lex_1), \gamma_i \in [0,1).
\label{eq:optimization}
\end{align}

This problem can be solved by \emph{alternating direction method of multipliers}~\cite{boyd2011distributed}.
The equality constraint can be incorporated into an augmented Lagrangian,
\begin{align}
\notag
L_\rho(\vg^{(0)},\vg^{(1)}) = & \frac{1}{2} \sum_{i\in \lex_0}(c_i - E[c_i])^2 + \frac{1}{2}\sum_{j \in \lex_1}(c_j - E[c_j])^2\\
& + \frac{\rho}{2}(\vmu^{(0)} \cdot \vg^{(0)} - \vmu^{(1)} \cdot \vg^{(1)})^2,
\end{align}
where $\rho > 0$ is the penalty parameter. The augmented Lagrangian is biconvex in $\vg^{(0)}$ and $\vg^{(1)}$, which suggests an iterative solution (Boyd et al. 2011, page 76)\nocite{boyd2011distributed}. Specifically, we hold $\vg^{(1)}$ fixed and solve for $\vg^{(0)}$, subject to $\gamma_i \in [0,1)$ for all $i \in \lex_0$. We then solve for $\vg^{(1)}$ under the same conditions. Finally, we update a dual variable $u$, representing the extent to which the equality constraint is violated. These updates are iterated until convergence. The unconstrained local updates to $\vg^{(0)}$ and $\vg^{(1)}$ can be computed by solving a system of linear equations, and the result can be projected back onto the feasible region. The penalty parameter $\rho$ is initialized at $1$, and then dynamically updated based on the primal and dual residuals (Boyd et al. 2011, pages 20-21)\nocite{boyd2011distributed}. 
\ifaaai
\else
More details are available in the appendix, and in the online source code.\footnote{\url{https://github.com/jacobeisenstein/probabilistic-lexicon-classification}}
\fi

\section{Evaluation}
An empirical evaluation is performed on four datasets in two languages. All datasets involve binary classification problems, and  performance is quantified by the \textbf{a}rea-\textbf{u}nder-the-\textbf{c}urve (AUC), a measure of classification performance that is robust to unbalanced class distributions. A perfect classifier achieves $\textsc{Auc}=1$; in expectation, a random decision rule gives $\textsc{Auc}=0.5$.

\paragraph{Datasets} The proposed method relies on co-occurrence counts, and therefore is best suited to documents containing at least a few sentences each. With this in mind, the following datasets are used in the evaluation:
\begin{description}
\item[Amazon] English-language product reviews across four domains; of these reviews, 8000 are labeled and another 19677 are unlabeled~\cite{blitzer2007biographies}.
\item[Cornell] 2000 English-language film reviews (version 2.0), labeled as positive or negative~\cite{pang2004sentimental}.
\item[CorpusCine] 3800 Spanish-language movie reviews, rated on a scale of one to five~\cite{vilares2015syntactic}. Ratings of four or five are considered as positive; ratings of one and two are considered as negative. Reviews with a rating of three are excluded.\item[IMDB] 50,000 English-language film reviews~\cite{maas2011learning}. This evaluation includes only the test set of 25,000 reviews, of which half are positive and half are negative.
\end{description}

\paragraph{Lexicons} Preliminary evaluation compared several English-language sentiment lexicons. The Liu lexicon~\cite{liu2015sentiment} 
consistently obtained the best performance on all three English-language datasets, so it was made the focus of all subsequent experiments. \citeauthor{ribeiro2016sentibench} (\citeyear{ribeiro2016sentibench}) also found that the Liu lexicon is one of the strongest lexicons for review analysis.
For the Spanish data, the ISOL lexicon was used~\cite{molina2013semantic}. It is a modified translation of the Liu lexicon.

\newcommand{\sysmult}{\sys{ProbLex-Mult}}
\newcommand{\sysdcm}{\sys{ProbLex-DCM}}

\paragraph{Classifiers} The evaluation compares the following unsupervised classification strategies:
\begin{description}
\item[\sys{Lexicon}] basic word counting, as in decision rule (\ref{eq:lexicon-decision-rule});
\item[\sys{Lex-Presence}] counting word presence rather than frequency, as in decision rule (\ref{eq:appearance-decision-rule});
\item[\sysmult] probabilistic lexicon-based classification, as proposed in this paper, using the multinomial likelihood model;
\item[\sysdcm] probabilistic lexicon-based classification, using the Dirichlet Compound Multinomial likelihood to reduce effective counts for repeated words;
\item[\sys{PMI}] An alternative approach, discussed in the related work, is to impute document labels from a seed set of words, and then compute ``sentiment scores'' for individual words from pointwise mutual information between the words and imputed labels~\cite{turney2002thumbs}. The implementation of this method is based on the description from \mycite{kiritchenko2014sentiment}, using the lexicons as the seed word sets.
\end{description}

As an upper bound, a supervised logistic regression classifier is also considered. This classifier is trained using five-fold cross validation. It is the only classifier with access to training data. For the \sysmult\ and \sysdcm\ methods, lexicon words which co-occur with the opposite lexicon at greater than chance frequency are eliminated from the lexicon in a preprocessing step.

\paragraph{Results}
Results are shown in Table~\ref{tab:results}. The superior performance of the logistic regression classifier confirms the principle that supervised classification is far more accurate than lexicon-based classification. Therefore, supervised classification should be preferred when labeled data is available. Nonetheless, the probabilistic lexicon-based classifiers developed in this paper (\sysmult\ and \sysdcm) go a considerable way towards closing the gap, with improvements in AUC ranging from less than 1\% on the CorpusCine data to nearly 8\% on the IMDB data. The \sys{PMI} approach performs poorly, improving over the simpler lexicon-based classifiers on only one of the four datasets. The word presence heuristic offers no consistent improvements, and the Bayesian adjustment to the classification rule (\sysdcm) offers only modest improvements on two of the four datasets.

\begin{table}
  \centering
  \begin{tabular}{lllll}
    \toprule
    & Amazon & Cornell & Cine & IMDB \\
    \midrule
    \sys{Lexicon} & .820 & .765 & .636 & .807\\
    \sys{Lex-Presence} & .820 & .770 &  .638 & .805 \\[3pt]
    \sys{PMI} & .793 & .761 & .638 & .868\\[3pt]
    \sysmult & .832 & .810 & \textbf{.644} & \textbf{.884}\\
    \sysdcm & \textbf{.836} & \textbf{.824} & \textbf{.645} & \textbf{.884} \\[3pt]
    \sys{LogReg} & .897 & .914 & .889 & .955 \\
   \bottomrule                                                     
  \end{tabular}
  \caption{Area-under-the-curve (AUC) for all classifiers. The best unsupervised result is shown in bold for each dataset.}
  \label{tab:results}
\end{table}


\section{Related work}
\mycite{turney2002thumbs} uses pointwise mutual information to estimate the ``semantic orientation'' of all vocabulary words from co-occurrence with a small seed set. This approach has later been extended to the social media domain by using emoticons as the seed set~\cite{kiritchenko2014sentiment}. Like the approach proposed here, the basic intuition is to leverage co-occurrence statistics to learn weights for individual words; however, PMI is a heuristic score that is not justified by a probabilistic model of the text classification problem. PMI-based classification underperforms \sysmult\ and \sysdcm\ on all four datasets in our evaluation.

The method-of-moments has become an increasingly popular estimator in unsupervised machine learning, with applications in topic models~\cite{anandkumar2014tensor}, sequence models~\cite{hsu2012spectral}, and more elaborate linguistic structures~\cite{cohen2014spectral}. Of particular relevance are ``anchor word'' techniques for learning latent topic models~\cite{Arora:12:Topic}. In these methods, each topic is defined first by a few keywords, which are assumed to be generated only from a single topic. From these anchor words and co-occurrence statistics, the topic-word probabilities can be recovered. A key difference is that the strong anchor word assumption is not required in this work: none of the words are assumed to be perfectly predictive of either label. We require only the much weaker assumption that words in a lexicon tend to co-occur less frequently with words in the opposite lexicon.

\section{Conclusion}
Lexicon-based classification is a popular heuristic that has not previously been analyzed from a machine learning perspective. This analysis yields two techniques for improving unsupervised binary classification: a method-of-moments estimator for word predictiveness, and a Bayesian adjustment for repeated counts of the same word. The method-of-moments estimator yields substantially better performance than conventional lexicon-based classification, without requiring any additional annotation effort. Future work will consider the generalization to multi-class classification, and more ambitiously, the extension to multiword units.

\vspace{-9pt}
\begin{small}
\paragraph{Acknowledgment} This research was supported by the National Institutes of Health under award number R01GM112697-01, and by the Air Force Office of Scientific Research.
\end{small}

\appendix
\setlength\parskip{6pt}

\section{Supplementary Material: Estimation Details}


This supplement describes the estimation procedure in more detail. The paper uses the method of moments to derive the following optimization problem,
\begin{align}
\notag
\min_{\vg^{(0)}, \vg^{(1)}} & \:\: \frac{1}{2} \sum_{i\in \lex_0}(c_i - E[c_i])^2 + \frac{1}{2}\sum_{j \in \lex_1}(c_j - E[c_j])^2\\
\notag
s.t. &  \:\:  \vmu^{(0)} \cdot \vg^{(0)} - \vmu^{(1)} \cdot \vg^{(1)}=0\\
\notag
& \forall_{\iinlex} 0 \leq \gamma^{(0)}_i < 1\\
& \forall_{\jinlex} 0 \leq \gamma^{(1)}_j < 1.
\label{eq:supp-optimization}
\end{align}
This problem is biconvex in the parameters $\vgx,\vgz$. We optimize using the \emph{alternating direction method of multipliers} (ADMM; Boyd et al. 2011)\nocite{boyd2011distributed}. In the remainder of this document, $x \cdot y$ is used to indicate a dot product between $x$ and $y$, and $x \odot y$ is used to indicate an elementwise product.

\subsection{ADMM for biconvex problems}
In general, suppose that the function $F(x,z)$ is biconvex in $x$ and $z$, and that the constraint $G(x,z) = 0$ is affine in $x$ and $z$,
\begin{align}
\min_{x,z} & F(x,z) \\
s.t.& G(x,z) = 0.
\end{align}

We can optimize via ADMM by the following updates (Boyd et al 2011, section 9.2),
\begin{align}
x^{k+1} \gets & \argmin{x} F(x,z) + (\rho/2)||G(x,z^k) + u^k||_2^2\\
z^{k+1} \gets & \argmin{z} F(x,z) + (\rho/2)||G(x^{k+1},z) + u^k||_2^2\\
u^{k+1} \gets & u^k + G(x^{k+1},z^{k+1}).
\end{align}

Now suppose we have a more general constrained optimization problem, 
\begin{align}
\min_{x,z} & \quad F(x,z) \\
\notag
s.t.& \quad G(x,z) = 0\\
\notag
& \quad x \in \set{C}_x\\
\notag
& \quad z \in \set{C}_z,
\end{align}
where $\set{C}_x$ and $\set{C}_z$ are convex sets. We can solve via the updates,
\begin{small}
\begin{align}
x^{k+1} \gets & \argmin{x \in \set{C}_x} F(x,z) + (\rho/2)||G(x,z^k) + u^k||_2^2\\
z^{k+1} \gets & \argmin{z \in \set{C}_z} F(x,z) + (\rho/2)||G(x^{k+1},z) + u^k||_2^2\\
u^{k+1} \gets & u^k + G(x^{k+1},z^{k+1}),
\end{align}
\end{small}
where $u$ is a \emph{dual variable} and $\rho > 0$ is a hyperparameter.

\subsection{Application to moment-matching}
In the application to moment-matching estimation, we have:
\begin{align}
x \triangleq & \vgx\\
z \triangleq & \vgz\\
G(x,z) \triangleq & \vmx \cdot \vgx - \vmz \cdot \vgz
\label{eq:equality-constraint}
\\
\set{C}_x = \set{C}_z \triangleq & [0,1)\\
F(x,z) \triangleq & \frac{1}{2}\sum_{i\in \lex_0}(c_i - E[c_i])^2 + \frac{1}{2}\sum_{\jinlex}(c_j - E[c_j])^2
\label{eq:objective}
\\
\notag
E[c_i] = & \sum_{\jinlex} E[c_{i,j}] =  s \mu_i \sum_{\jinlex} \mu_j (1-\gamma_i\gamma_j) \\
= & s \mu_i \sum_{\jinlex} \mu_j - s \mu_i \gamma_i \sum_{\jinlex} \mu_j \gamma_j\\
\notag
E[c_j] = & \sum_{\iinlex} E[c_{i,j}] = s \mu_j \sum_{\iinlex} \mu_i (1 - \gamma_i\gamma_j)\\
 = & s\mu_j \sum_{\iinlex} \mu_i - s \mu_j \gamma_j \sum_{\iinlex} \mu_i \gamma_i \\
s \triangleq & \sum_t N_t(N_t-1).
\end{align}

We now consider how to perform the updates to $x^{k+1}$ as a quadratic closed-form expression (an identical derivation applies to $z^{k+1}$). Specifically, if the overall objective for $x$ can be written in the form,
\begin{align}
J(x) = & \frac{1}{2}x^T P x + q \cdot x + r,
\end{align}
then the optimal value of $x$ is found at,
\begin{align}
\hat{x} = & -P^{-1} q.
\end{align}
We will obtain this form by converting the objective $F$ and the tersm relating to the equality constraint $G$ the boundary constraint $H$ into quadratic forms.

\subsubsection{Objective}
We define helper notation,
\begin{align}
r_i = & c_i - s\mu_i \sum_{\jinlex}\mu_j\\
r_j = & c_j - s\mu_j \sum_{\iinlex}\mu_i,
\end{align}
representing the \emph{residuals} from a model in which $\gamma_i = \gamma_j = 0$ for all $i$ and $j$. Using these residuals, we rewrite the objective from \autoref{eq:objective},
\begin{small}
\begin{align}
F(\vgx,\vgz) = & \frac{1}{2}\sum_{i\in \lex_0}(c_i - E[c_i])^2 + \frac{1}{2}\sum_{\jinlex}(c_j - E[c_j])^2\\
\notag = &\frac{1}{2}\sum_{\iinlex}(r_i + s \mu_i \left(\sum_{\jinlex} \mu_j \gamma_j \right) \gamma_i )^2\\
& + \frac{1}{2}\sum_{\jinlex}(r_j + s \mu_j \left(\sum_{\iinlex} \mu_i \gamma_i \right) \gamma_j )^2.
\end{align}
\end{small}

Solving first for $\vgx$, we can rewrite the left term as a quadratic function,
\begin{small}
\begin{align}
\frac{1}{2}\sum_{\iinlex}(c_i - E[c_i])^2 = & \frac{1}{2} (\vgx)^T P_0 \vgx + \vq_0 \cdot \vgx + \frac{1}{2} \sum_{\iinlex}r^2_i\\
(P_0)_{ii} = & (s^2 (\sum_{\jinlex} \mu_j \gamma_j ) \mu_i^2) \\
(P_0)_{i\neq j}=& 0 \\
\vq_0 = & s (\sum_{\jinlex} \mu_j \gamma_j) (\vec{r}^{(0)} \odot \vmx),
\end{align}
\end{small}
where the matrix $P_0$ is diagonal. We can also rewrite the second term as a quadratic function of $\vgx$,
\begin{small}\begin{align}
\frac{1}{2}\sum_{\jinlex}(c_j - E[c_j])^2 = & \frac{1}{2} (\vgx)^T P_1 \vgx + \vq_1 \cdot \vgx + \frac{1}{2} \sum_{\jinlex}r^2_j\\
P_1 = & s^2 (\sum_j \mu_j^2 \gamma_j^2) \vmx (\vmx)^T\\
\vq_1 = & s (\sum_j r_j \mu_j \gamma_j) \vmx,
\end{align}
\end{small}
where the matrix $P_1$ is rank one. To summarize the terms from the objective,
\begin{small}
\begin{align}
\notag
P_F = & \text{Diag}\left(s^2 \left[ \sum_{\jinlex} \mu_j \gamma_j \right] \vmx \odot \vmx\right)\\
& + s^2 (\sum_{\jinlex} \mu_j^2 \gamma_j^2) \vmx (\vmx)^T\\
q_F = & s (\sum_{\jinlex} \mu_j \gamma_j) (\vec{r}^{(0)} \odot \vmx) + s (\sum_{\jinlex} r_j \mu_j \gamma_j) \vmx
\end{align}
\end{small}
We get an analogous set of terms when solving for $\vgz$, meaning that we can use the same code, with a change over arguments.

\subsubsection{Equality constraint}
The constraint $G$ requires that equal weight be assigned to the two lexicons,
\begin{align}
G(\vgx,\vgz) = & \vmx \cdot \vgx - \vmz \cdot \vgz
\end{align}
Thus, the augmented Lagrangian term $(\rho/2)||G(\vgx,\vgz) + u^k||_2^2$ can be written as a quadratic function of $\vgx$,
\begin{align}
\notag
& (\rho/2)||G(\vgx,\vgz) + u^k||_2^2\\
& = (\rho/2)(\vmx \cdot \vgx - \vmz \cdot \vgz + u^k)^2\\
& = \frac{1}{2} (\vgx)^T P_G \vgx + \vq_G \cdot \vgx + \ldots.
\end{align}
This quadratic form for $\vgx$ has the parameters,
\begin{align}
P^{(0)}_G = & \rho \vmx (\vmx)^T\\
q^{(0)}_G = & \rho (u^k - \vmz \cdot \vgz ) \vmx.
\end{align}

When solving for $\vgz$, we have,
\begin{align}
& (\rho/2)||G(\vgx,\vgz) + u^k||_2^2\\
& = (\rho/2)(\vmx \cdot \vgx - \vmz \cdot \vgz + u^k)^2\\
& = \frac{1}{2} (\vgz)^T P^{(1)}_G \vgz + \vq^{(1)}_G \cdot \vgz + \ldots,
\end{align}
so that,
\begin{align}
P^{(1)}_G = & \rho \vmz (\vmz)^T\\
q^{(1)}_G = & -\rho(u^k + \vmx \cdot \vgx)\vmz\\
= & \rho(-u^k - \vmx \cdot \vgx)\vmz,
\end{align}
meaning that we can use the same code, but plug in $-u^k$ instead of $u^k$.

\newcommand{\Px}{\ensuremath P^{(0)}}
\newcommand{\qx}{\ensuremath q^{(0)}}
\newcommand{\Pz}{\ensuremath P^{(1)}}
\newcommand{\qz}{\ensuremath q^{(1)}}

\subsubsection{Unconstrained solution}
The augmented Lagrangian for $\vgx$ can be written as,
\begin{align}
J(\vgx) = & \frac{1}{2} (\vgx)^T \Px \vgx + \qx \cdot \vgx + r
\label{eq:augmented-lagrangian}
\\
\Px = & \Px_{\text{diag}} + \Px_{\text{low-rank}}\\
\Px_{\text{Diag}} = & \text{Diag}\left(s^2 \left[ \sum_{\jinlex} \mu_j \gamma_j \right] \vmx \odot \vmx\right)\\
\Px_{\text{Low-rank}} = & (s^2 (\sum_{\jinlex} \mu_j^2 \gamma_j^2)  + \rho) \vmx (\vmx)^T \\
\notag
\qx = & s (\sum_{\jinlex} \mu_j \gamma_j) (\vec{r}^{(0)} \odot \vmx) \\
\notag  & + s (\sum_{\jinlex} r_j \mu_j \gamma_j) \vmx \\
& + \rho (u^k - \vmz \cdot \vgz ) \vmx
\end{align}
Ignoring the constraint set $\set{C}_x$, the solution for $\vgx$ is given by,
\begin{align}
\vg^{(0)} \gets & -(\Px_{\text{diag}} + \Px_{\text{low-rank}})^{-1} \qx.
\label{eq:unconstrained-solution}
\end{align}
The solution can be computed using the Woodbury identity.

\begin{algorithm}[t!]
\begin{algorithmic}
\While{global primal and dual residuals are above threshold}
\State{$\Px,\qx \gets \text{ComputeQuadraticForm}(\vgx,\vgz,u)$}
\State{$a \gets 0, v \gets 0$}
\While{local primal and dual residuals are above threshold}
\State{$\vgx \gets (\Px + \rho_2 \mathbb{I})^{-1}(\qx + \rho_2 (v - a))$}
\State{$a \gets \Pi_{\set{C}_x}(\vgx + v)$}
\State{$v \gets v+ \vgx - a$}
\EndWhile
\State{$\vgx \gets \Pi_{\set{C}_x}(\vgx)$}
\State{$\Pz,\qz \gets \text{ComputeQuadraticForm}(\vgz,\vgx,-u)$}
\State{$b \gets 0, w \gets 0$}
\While{local primal and dual residuals are above threshold}
\State{$\vgz \gets (\Pz + \rho_2 \mathbb{I})^{-1}(\qz + \rho_2 (w - b))$}
\State{$b \gets \Pi_{\set{C}_z}(\vgz + w)$}
\State{$w \gets w + \vgz - b$}
\EndWhile
\State{$\vgz \gets \Pi_{\set{C}_z}(\vgz)$}
\State{$u \gets u + \vmx \cdot \vgx - \vmz \cdot \vgz$}
\EndWhile
\end{algorithmic}
\caption{ADMM optimization for unsupervised lexicon-based classification}
\label{alg:optimization}
\end{algorithm}

\subsubsection{Constrained solution}
Each update to $\vgx$ and $\vgz$ must lie within the constraint sets $\set{C}_x$ and $\set{C}_z$. One way to ensure this is to apply boundary-constrained L-BFGS to the augmented Lagrangian in \autoref{eq:augmented-lagrangian}. This solution requires the gradient, which is simply $P \vgx + \qx$. A slightly faster (and more general) solution is to apply ADMM again, using the following iterative updates (Boyd et al. 2011, page 33):
\begin{small}
\begin{align}
\vgx \gets & -(\Px_{\text{diag}} + \Px_{\text{low-rank}} + \rho_2 \mathbb{I})^{-1} (\qx + \rho_2(v-a))\\
a \gets & \Pi_{\set{C}_x}(\vgx)\\
v \gets & v + \vgx - a,
\end{align}
\end{small}
where $\Pi_{\set{C}_x}$ projects on to the set $\set{C}_x$, and $v$ is an additional dual variable. This requires only a minor change to the quadratic solution in \autoref{eq:unconstrained-solution}: we add $\rho_2$ to the diagonal of $P$, and we add $\rho_2 (v - a)$ to the vector $q$. 

The overall algorithm is listed in \autoref{alg:optimization}. Each loop terminates when the primal and dual residuals fall below a threshold (Boyd et al 2011, pages 19-20). We also use these residuals to dynamically adapt the penalties $\rho$ and $\rho_2$ (Boyd et al 2011, pages 20-21).

\bibliographystyle{aaai}
\bibliography{cite-strings,cites,cite-definitions}

\end{document}